\documentclass[runningheads]{llncs}
\usepackage[T1]{fontenc}
\usepackage{graphicx}
\usepackage{amsmath}
\usepackage{amsfonts}
\usepackage{amssymb}
\usepackage{algorithm}
\usepackage{algpseudocode}
\usepackage{graphicx} 
\usepackage{booktabs} 
\usepackage{multirow}
\usepackage[table]{xcolor}

\usepackage{marvosym,graphicx} 
\newcommand{\corrauthor}{\textsuperscript{\scalebox{0.85}{\Letter}}} 
\usepackage{hyperref}
\usepackage{rotating}
\usepackage{graphicx}
\begin{document}
\title{Model Merging to Evolution: Parameter Space Exploration for Expert Models}
\titlerunning{Model Merging to Evolution}
%
\author{Chao Wang\inst{1}\textsuperscript{\dag}\corrauthor \and
Yuchen Guo\inst{1}\textsuperscript{\dag} \and Zheng Tan\inst{2,3} \and Guanchun Wang\inst{1} \\ \and Yanbiao Ma\inst{4}  \and Qiqi Duan\inst{5} \and Peng Wu\inst{6}}
\authorrunning{C. Wang et al.}
%
\institute{School of Artificial Intelligence, Xidian University, Xi'an, China
\email{xiaofengxd@126.com} \and 
School of Computer Science, University of Birmingham, Birmingham, UK \and
Department of Computer Science and Engineering, Southern University of Science and Technology, Shenzhen, China \and
Gaoling School of Artificial Intelligence, Renmin University of China, Beijing, China \and
School of Computing and Artificial Intelligence, Jiangxi University of Finance and Economics, Nanchang, China \and
School of Computer Science, Northwestern Polytechnical University, Xi'an, China}
\maketitle              
\begingroup
\renewcommand\thefootnote{\dag}
\footnotetext{Equal contribution.}
\endgroup

\begin{abstract}
Model merging integrates the capabilities of multiple expert models to create strong models for multiple tasks without additional training, thereby reducing computational resource requirements. However, existing methods operate within the convex combination space of expert models, failing to explore high-performance regions outside this space. This paper proposes the MERGEvolve framework, which unifies model merging and evolution within an evolution strategy by treating the merged model as the initialization for evolutionary exploration of the parameter space. During the merging phase, expert models act as deterministic sources to build a strong initial point. The evolution phase then explores the parameter space using random noise. Theoretical analysis shows that MERGEvolve explores regions outside the convex combination space. Extensive experiments on single-task and multi-task benchmarks demonstrate that MERGEvolve consistently achieves performance competitive with advanced model merging baselines. Ablation studies confirm that a high-quality initial point is critical for efficient exploration of the parameter space.

\end{abstract}
\section{Introduction}

The rapid development of large language models (LLMs) has fostered numerous expert models that perform well on specific tasks \cite{NEURIPS2020_1457c0d6}. However, maintaining many models incurs significant computational and storage overhead \cite{pmlr-v162-wortsman22a}. Model merging techniques address this limitation by integrating the capabilities of multiple expert models into a single model. The integration reduces training resource consumption while enabling diverse combinations of model capabilities \cite{10.1145/3787849}. However, as the number of merged models increases, performance gains saturate quickly or even decline \cite{wang2025expertsfailtheoreticalanalysis}. This phenomenon limits the applicability of model merging in scenarios that require integrating numerous expert models.

The collaborative integration of multiple expert models primarily involves static model merging and model evolution. Static model merging methods linearly combine the parameters of multiple models to construct a unified model based on arithmetic rules \cite{10.1145/3787849,wang2025expertsfailtheoreticalanalysis}. For instance, Task Arithmetic \cite{ilharco2023editing} distinguishes model importance using predefined weights, while Fisher Merging~\cite{matena2022merging} derives the combination weights from the Fisher information matrix. TIES~\cite{yadav2023ties} further mitigates parameter interference by trimming parameters and selecting signs. Although static methods succeed in merging a small number of expert models, the search space remains confined to convex combination regions. Static methods struggle to explore nonlinear areas within the parameter space.

Model evolution provides a new perspective for expert model collaboration \cite{minut-etal-2025-mergenetic}. Akiba et al. \cite{2025Evolutionary} employ the covariance matrix adaptation evolution strategy (CMA-ES) to automatically configure the convex combination weights for model merging. Model Swarms \cite{feng2025model} uses particle swarm optimization (PSO) to conduct collaborative searches in the parameter space by linearly combining expert models. However, Model evolution relies primarily on heuristic operations and lacks a theoretical connection with model merging, resulting in limited interpretability.

\begin{figure}
\includegraphics[width=\textwidth]{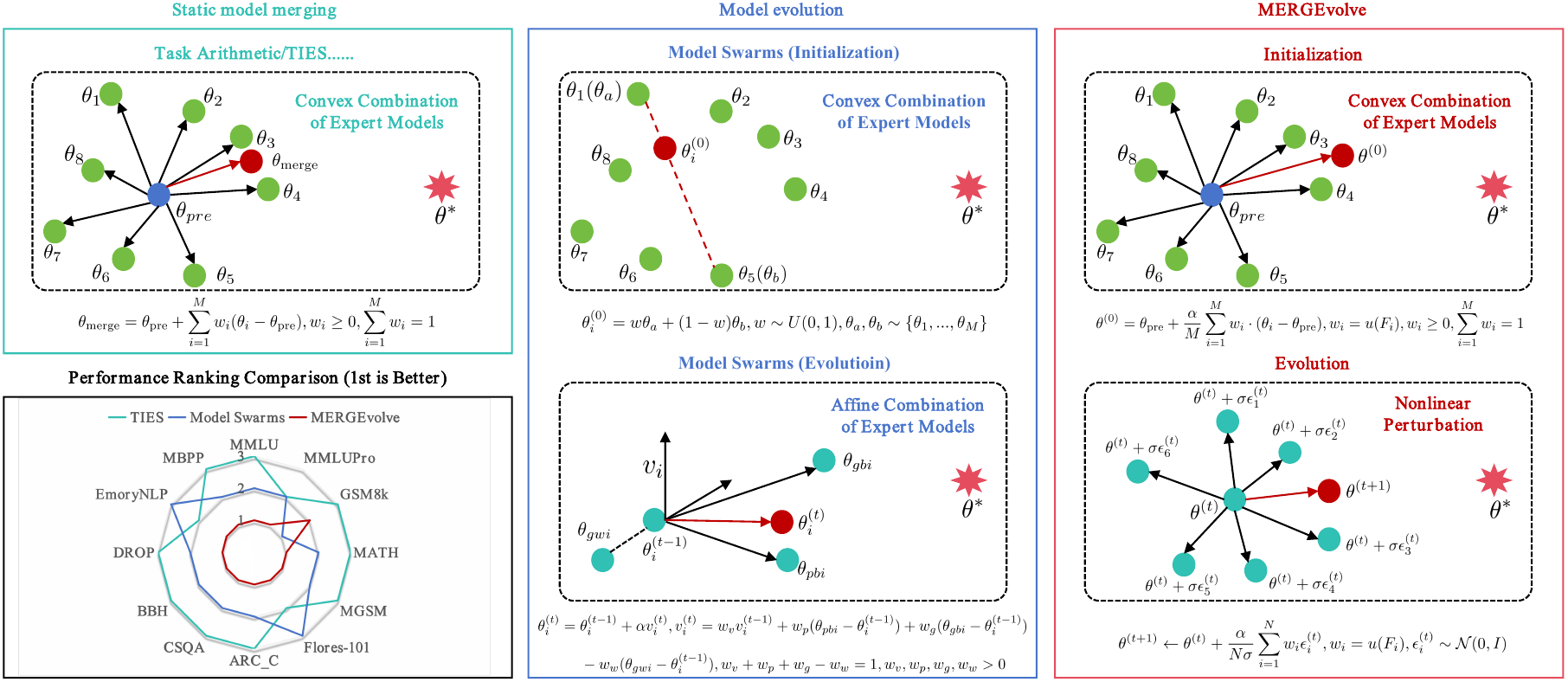}
\caption{Concept: Static model merging and model evolution focus on convex and affine combinations of experts, respectively, whereas our unified ES framework, MERGEvolve, treats model merging as initialization for evolution. Performance: Rankings of representative methods (TIES, Model Swarms, MERGEvolve) on 12 single-task datasets (1st is best). See the Experiments section for comprehensive results.} \label{fig1}
\end{figure}

To address these challenges, we establish a unified evolution strategy (ES) framework, called MERGEvolve, to integrate model merging with model evolution. As shown in Fig.~\ref{fig1}, MERGEvolve treats the merged model as the initialization for parameter space exploration. The merging phase utilizes expert models as deterministic sources to construct a high-quality initial point. The evolution phase explores the parameter space through random noise. The two phases share unified update rules and differ only in the perturbation sources. Perturbations in the merging phase originate from vectors $(\theta_i-\theta_{\text{pre}})$ between the expert models $\theta_i$ and the base pre-trained model $\theta_{\text{pre}}$. Perturbations in the evolution phase originate from random noise $\epsilon\sim\mathcal{N}(0,I)$. The theoretical consistency ensures a smooth transition from merging to evolution. Theoretical analysis shows that MERGEvolve overcomes the convex combination space limitation and explores regions beyond the restricted space. Extensive experiments on multiple datasets indicate that MERGEvolve achieves competitive performance compared to advanced model merging baselines. Ablation studies further verify the critical role of a high-quality initial point for parameter space exploration.

\section{Related Work}
Model merging has emerged as a computationally efficient paradigm to synthesize the capabilities of multiple expert LLMs without additional training. Recent advancements can be broadly categorized into two primary methods: static model merging and model evolution.

\textbf{Static Model Merging:} Static methods construct a unified model by linearly combining the parameters of expert models based on fixed rules or arithmetic operations. Task Arithmetic \cite{ilharco2023editing} manipulates model capabilities by directly adding or subtracting task vectors ($\tau_i = \theta_i - \theta_{\text{pre}}$). Parameter importance was introduced by leveraging the Fisher information matrix to identify task-critical weights \cite{matena2022merging}, subsequently facilitating model merging via importance-aware weighted averaging. Similarly, TIES \cite{yadav2023ties} was proposed to mitigate parameter interference through a resolve-and-merge strategy. However, TIES's efficacy diminishes substantially as the number of integrated models scales upward \cite{wang2025expertsfailtheoreticalanalysis}.

\textbf{Model Evolution:} With the rapid development of model merging technology, integrating evolutionary algorithms (EAs) \cite{10,hansen2016cma,yan2024populating,chao2024match} into the model merging is becoming increasingly popular. It is worth noting that Akiba et al. \cite{2025Evolutionary} divided the model merging process into two distinct configuration domains: the parameter space and the data flow space. Within this paradigm, ``merging recipes'' are autonomously identified through evolutionary search. By employing CMA-ES to optimize the convex combination weights of the constituent models, this approach shifts the optimization focus toward the design of merging strategies, thereby addressing the question of ``how to merge''. Model Swarms \cite{feng2025model} conceptualizes LLM experts as ``particles'', leveraging PSO to iteratively combine them via linear weighting in the parameter space. Subsequently, Zhang et al. \cite{zhang2025nature} introduced evolutionary operators, including crossover, mutation, and selection, to explore the parameter space. However, these approaches either remain heavily reliant on heuristics or lack an interpretable theoretical connection between model evolution and model merging.

\section{MERGEvolve Framework}
\label{sec4}
Our MERGEvolve framework is built on a key insight: model merging should not be the culmination of parameter space exploration, but rather as its point of departure. We present an overview of MERGEvolve in \textbf{Algorithm \ref{alg:merge_evolution}}. Given the pre-trained model $\theta_{\text{pre}}$ and a set of $M$ expert models $\{\theta_i\}_{i=1}^M$, our goal is to identify the optimal parameters $\theta^*$ that maximize the model's performance across target tasks.

\begin{algorithm}[htbp]
\caption{MERGEvolve}
\label{alg:merge_evolution}
\begin{algorithmic}[1]
\Require Pre-trained model $\theta_{\text{pre}}$; Expert models $\{\theta_i\}_{i=1}^{M}$; Downstream target task $F$; Utility function $u$; Learning rate $\alpha$; Number of iterations $T$; Population size $N$; Standard deviation $\sigma$.
\Ensure Model $\theta^{T}$.

\Statex \textit{Merging phase: Constructing a high-quality initial point}
\For{$i = 1$ to $M$}
    \State $F_i \gets \mathrm{Performance}(\theta_i, F)$;
\EndFor
\State sort {$(\theta_i, F_i)$} with respect to $F_i$ and $w_i \gets u(F_i), i=1,...,M$;
\State $\theta^{(0)} \gets \theta_{\text{pre}} + \frac{\alpha}{M} \sum_{i=1}^{M} w_i (\theta_i - \theta_{\text{pre}})$;

\Statex \textit{Evolutionary phase: Parameter-space exploration}
\For{$t = 0$ to $T-1$}
    \State Sample perturbations $\epsilon_1^{(t)}, \epsilon_2^{(t)}, \ldots, \epsilon_{N}^{(t)} \sim \mathcal{N}(0, I)$;
    \For{$i = 1$ to $N$}
        \State $F_i \gets \mathrm{Performance}(\theta^{(t)} + \sigma \epsilon_i^{(t)}, F)$;
    \EndFor
    \State sort {$(\epsilon_i^{(t)}, F_i)$} with respect to $F_i$ and $w_i \gets u(F_i), i=1,...,N$;
    \State $\theta^{(t+1)} \gets \theta^{(t)} + \frac{\alpha}{N \sigma} \sum_{i=1}^{N} w_i \epsilon_i^{(t)}$;
\EndFor
\State \Return $\theta^{T}$.
\end{algorithmic}
\end{algorithm}

\subsection{Merging Phase: Constructing High-Quality Initializations}
The merging phase can be conceptualized as a deterministic approximation of an ES \cite{hansen2016cma,salimans2017evolution,wierstra2014natural} applied over the discrete distribution formed by expert models. In this stage, the parameters are updated as follows:
\begin{equation}
\theta^{(0)} = \theta_{\text{pre}} + \frac{\alpha}{M} \sum_{i=1}^{M} w_i \cdot (\theta_i - \theta_{\text{pre}}),
\end{equation}
where $\theta_i$ represents the $i$-th expert model, with $\theta_i - \theta_{\text{pre}}$ denoting its parameter offset relative to the pre-trained model. $w_i$ is a utility weight derived from the evaluation performance of each expert on downstream target tasks. $\alpha$ is a scaling coefficient that governs the merging intensity. Unlike heuristic-based merging methods \cite{feng2025model}, this approach offers a clear theoretical foundation through the perspective of ES.

\subsection{Evolutionary Phase: Exploration in Non-Convex Spaces}
The evolutionary phase performs continuous optimization starting from the merged initialization point. The update formula is defined as:
\begin{equation}
\theta^{(t+1)} = \theta^{(t)} + \frac{\alpha}{N \sigma} \sum_{i=1}^{N} w_i \cdot \epsilon_i^{(t)},
\end{equation}
where $\sigma$ is the noise standard deviation and $\epsilon_i^{(t)} \sim \mathcal{N}(0, I)$ represents stochastic noise sampled from a standard normal distribution. $w_i$ is a utility weight $u(F_i)$ derived from the evaluation results of each model $\{\theta^{(t)} + \sigma \epsilon_i^{(t)}\}$ on downstream target tasks.

\section{Theoretical Analysis}
\label{sec2}
\label{sec3}
This section provides a theoretical analysis of popular model merging methods and demonstrates key properties regarding their capacity for parameter space exploration. We demonstrate that static model merging is confined to the convex hull formed by the initial experts, thereby restricting its search space to this bounded region (\textbf{Theorem 1}). While model evolution's methods, such as Model Swarms, transcend the boundaries of the initial convex hull, they remain inherently restricted to the finite affine hull spanned by the original experts (\textbf{Section \ref{sec3_3}}). Using these theoretical insights, we establish that MERGEvolve overcomes convex hull constraints (\textbf{Theorem 5}), enabling access to latent regions that are fundamentally unreachable by conventional paradigms.

\subsection{Static Model Merging}

\textbf{Definition 1}(Static Model Merging). Given a set of experts $\{\theta_1,\theta_2,\dots,\theta_M\}$ is fine-tuned from a base pre-trained model $\theta_{\text{pre}}$. Static model merging constructs a unified model $\theta_{merge}$ through a weighted combination of the experts:

\begin{equation}
\theta_{\text{merge}} = \theta_{\text{pre}} + \sum_{i=1}^{M} w_i (\theta_i - \theta_{\text{pre}}),
\end{equation}
where $w_i \in \mathbb{R}$ denotes the merging weight (coefficient) that determines the contribution of each expert. In the context of parameter-efficient fine-tuning (PEFT), such as low-rank adaptation (LoRA) \cite{hu2022lora}, expert models are represented as low-rank adaptations, and the merging process operates directly on these adaptive parameters. Let $\{\Delta\theta_1, \Delta\theta_2, \dots, \Delta\theta_M\}$ represent the parameter deviations from the pre-trained model (i.e., $\Delta\theta_i = \theta_i - \theta_{\text{pre}}$). Consequently, static model merging can be equivalently formulated as:

\begin{equation}
\theta_{\text{merge}} = \theta_{\text{pre}} + \sum_{i=1}^{M} w_i \Delta\theta_i.
\end{equation}
This formulation captures the underlying principles of prevalent static merging techniques, such as Task Arithmetic, Fisher Merging, and TIES-MERGING.

\subsection{Convex Combination in Static Model Merging}
\label{sec3_2}
\textbf{Theorem 1} (Convex Combination in Static Model Merging). When the merging coefficients satisfy $\sum_{i=1}^M w_i = 1$ and $w_i \geq 0$ for all $i$, the merged model in static model merging represents a convex combination of the experts within the parameter space.

\begin{proof}
Assume the merged model is parameterized by:
\begin{equation}
    \theta_{\text{merge}}=\theta_{\text{pre}}+\sum_{i=1}^{M}w_i(\theta_i-\theta_{\text{pre}})=\left(1-\sum_{i=1}^{M}w_i\right)\theta_{\text{pre}}+\sum_{i=1}^{M}w_i\theta_i. 
\end{equation}
  By applying the constraints $\sum_{i=1}^{M}w_i=1$ and $w_i\geq0$, we have $\theta_{\text{merge}}=\sum_{i=1}^{M}w_i\theta_i$, which constitutes a convex combination of the experts. 
This confirms that static model merging under coefficient constraints operates only within the convex hull of the experts, thus restricting parameter exploration to the convex combination space.
\end{proof}

\subsection{Model Swarms: Convex Combination in Initialization and Affine Combination in Evolution}
\label{sec3_3}
\textbf{Theorem 2} (Convex Combination in Initialization). In the initialization phase of Model Swarms, the initial particles are constructed as a convex combination of expert models.

\begin{proof}
The initialization phase expands the initial expert pool through pairwise crossover and linear interpolation \cite{feng2025model}. Specifically, two experts, $\theta_a$ and $\theta_b$, are randomly selected from the initial experts $\{\theta_1, \dots, \theta_M\}$. By sampling a weight $w$ from a uniform distribution $U(0, 1)$, a new particle is obtained by $\theta_i^{(0)} = w\theta_a + (1 - w)\theta_b$, where $w \in [0, 1]$ and $w + (1 - w) = 1$. By iterating this procedure $N - M$ times, the initial experts $\{\theta_i\}_{i=1}^M$ is expanded to a larger population $\{\theta_i\}_{i=1}^N$. For each new particle $\theta_i^{(0)}$, we have:

\begin{equation}
\theta_i^{(0)} = w\theta_a + (1 - w)\theta_b, \quad w \sim U(0, 1), \quad \theta_a, \theta_b \in \{\theta_1, \dots, \theta_M\}.
\end{equation}
Given that $w \in [0, 1]$, all initial particles are situated within the convex hull of the original experts.
\end{proof}

\textbf{Theorem 3} (Affine Combination in Evolution). The evolution phase of Model Swarms updates particle positions through affine combinations.

\begin{proof}
The evolution phase adheres to the following update dynamics: 
\begin{equation}
\begin{aligned}
& v_i^{(t)} = w_v v_i^{(t-1)} + w_p (\theta_{pbi} - \theta_i^{(t-1)}) + w_g (\theta_{gbi} - \theta_i^{(t-1)}) - w_w (\theta_{gwi} - \theta_i^{(t-1)}), \\
& \theta_i^{(t)} = \theta_i^{(t-1)} + \alpha v_i^{(t)},
\end{aligned}
\end{equation}
where $\theta_{pbi}, \theta_{gbi}, \theta_{gwi}$ denote the personal best, global best, and global worst positions up to iteration $t-1$. The coefficients satisfy $w_v, w_p, w_g, w_w > 0, w_v + w_p + w_g - w_w = 1$ and $\alpha > 0$.

We prove by induction that for all $t \ge 1$, $\theta_i^{(t)}$ is an affine combination of the initial particles $\{\theta_j^{(0)}\}_{j=1}^N$. (The case $t=0$ holds trivially by definition.)

\textbf{Base Case} ($t=1$): 
At $t=1$, $\theta_i^{(1)} = \theta_i^{(0)} + \alpha v_i^{(1)}$. Since $v_i^{(1)}$ is a linear combination of $\theta_{pbi}, \theta_{gbi}, \theta_{gwi}$ (all equal to initial positions at $t=0$) and $\theta_i^{(0)}$, and the coefficients in the update rule sum to $1$, $\theta_i^{(1)}$ is an affine combination of $\{\theta_j^{(0)}\}_{j=1}^N$.

\textbf{Inductive Step}: 
Assume that for all $s < t$, every $\theta_k^{(s)}$ is an affine combination of $\{\theta_j^{(0)}\}_{j=1}^N$. From the position update at $t-1$, we have $v_i^{(t-1)} = \frac{1}{\alpha}(\theta_i^{(t-1)} - \theta_i^{(t-2)})$. Substituting this into the update for $\theta_i^{(t)}$ yields:
\begin{equation}
\begin{aligned}
\theta_i^{(t)} ={}& (1 + w_v - \alpha w_p - \alpha w_g + \alpha w_w) \theta_i^{(t-1)} - w_v \theta_i^{(t-2)} \\
&+ \alpha w_p \theta_{pbi} + \alpha w_g \theta_{gbi} - \alpha w_w \theta_{gwi}.
\end{aligned}
\end{equation}
Summing the coefficients of all terms, we have:
\begin{equation}
(1 + w_v - \alpha w_p - \alpha w_g + \alpha w_w) + (-w_v) + \alpha w_p + \alpha w_g - \alpha w_w = 1.
\end{equation}
Since the affine hull of $\{\theta_j^{(0)}\}_{j=1}^N$ is closed under linear combinations with coefficient sum 1, $\theta_i^{(t)}$ remains an affine combination of the initial particles. Notably, the negative coefficient $-\alpha w_w$ on $\theta_{gwi}$ makes this an affine (rather than convex) combination, allowing exploration beyond the initial convex hull while staying within the affine hull.
\end{proof}

\subsection{MERGEvolve: Convex Combination in Initialization and Perturbations in Evolution}
\label{sec3_4}
\textbf{Theorem 4} (Convex Combination in Initialization). During the initialization phase of MERGEvolve, convex combinations of expert models are constructed as a high-quality initial point.

Proof. In MERGEvolve, the initialization phase adheres to the principles of static model merging, i.e., $\theta^{(0)} = \theta_{\text{pre}} + \frac{\alpha}{M}\sum_{i=1}^{M} w_i (\theta_i - \theta_{\text{pre}})$, where the merging coefficients $w_i$ satisfy the constraints $\sum_{i=1}^{M} w_i = 1$ and $w_i \geq 0$ for all $i$. 

As shown in \textbf{Theorem 1}, when $\alpha = m$, we have $\theta^{(0))} = \sum_{i=1}^{M} w_i \theta_i$, which constitutes a convex combination of the expert models. This initialization strategy provides a high-quality starting point within the convex hull of the expert models, effectively leveraging the complementary capabilities of diverse experts.

\textbf{Theorem 5} (Perturbations in Evolution). MERGEvolve explores the parameter space through perturbations applied to the merged model in the evolution phase, extending the search trajectory beyond the space of convex combinations.

Proof. In MERGEvolve, the evolution phase adheres to the update dynamics $\theta^{(t+1)} = \theta^{(t)} + \frac{\alpha}{N \sigma} \sum_{i=1}^{N} w_i \cdot \epsilon_i^{(t)}$. Beginning from the initialization point $\theta^{\text{(0)}}$, we obtain:
\begin{equation}
\theta^{(T)} = \sum_{i=1}^{M} w_i \theta_i \;+\; \frac{\alpha}{N \sigma} \sum_{t=0}^{T-1} \sum_{i=1}^{N} w_i \, \epsilon_i^{(t)}.
\end{equation}
The final model represents the sum of a convex combination of initial experts and a series of random perturbations. Unlike Model Swarms, MERGEvolve is not confined to the affine hull of the initial experts.

\section{Experiments}
\label{sec5}
This section evaluates MERGEvolve across single-task and multi-task settings, assesses its zero-shot generalization to unseen tasks, and validates the contribution of each component through ablation studies. Furthermore, we analyze the emergent capabilities of evolved models and investigate the impact of expert diversity on the performance of MERGEvolve.

\subsection{Experimental Setup}
\label{sec5_1}

\subsubsection{Benchmarks}
To comprehensively evaluate MERGEvolve, we curate a diverse benchmark suite comprising widely-adopted datasets, designed to probe seven distinct cognitive dimensions of LLMs rather than focusing on a single domain:
\begin{sloppypar}
\begin{itemize}
    \item \textbf{Reasoning and Mathematics:} Logical reasoning is evaluated on DROP and BBH \cite{dua2019drop,suzgun2023challenging}, while mathematical problem-solving is assessed via MATH, GSM8K, and MGSM \cite{hendrycks2021measuring,cobbe2021training,shi2022language}.
    \item \textbf{Knowledge and QA:} General knowledge and question answering are tested using MMLU, MMLU-Pro, ARC\_C, and CSQA \cite{hendrycks2020measuring,wang2024mmlu,clark2018think,talmor2019commonsenseqa}.
    \item \textbf{Specialized Domains:} Code generation, affective computing, and multilingual processing are evaluated on MBPP \cite{austin2021program}, EmoryNLP and MELD \cite{zahiri2018emotion,poria2019meld}, and Flores-101 \cite{goyal2022flores}, respectively.
\end{itemize}
\end{sloppypar}

\begin{table}[htbp]
\centering
\caption{Detailed information of the datasets.}
\small
\label{tab:datasets}
\begin{tabular}{lllcc}
\hline
 \multirow{2}{*}{\textbf{Dataset}} & \multirow{2}{*}{\textbf{Category}} & \multirow{2}{*}{\textbf{Metric}} & \multicolumn{2}{c}{\textbf{Size}} \\ \cline{4-5} 
  &  &  & \textbf{validation} & \textbf{test} \\ \hline
DROP & Logical Reasoning & EM & 200 & 1000 \\
BBH & Logical Reasoning & Accuracy & 200 & 1000 \\
MATH & Mathematics & Accuracy & 200 & 1000 \\
GSM8k & Mathematics & Accuracy & 200 & 1000 \\
MGSM & Multilingual Processing, Mathematics & Accuracy & 200 & 2637 \\
MMLU & General Knowledge & Accuracy & 200 & 1000 \\
MMLUPro & General Knowledge & Accuracy & 200 & 1000 \\
ARC\_C & Question Answering & Accuracy & 200 & 1000 \\
CSQA & Question Answering & Accuracy & 200 & 1000 \\
MBPP & Code Generation & Pass@1 & 200 & 774 \\
EmoryNLP & Affective Computing & Weighted F1 & 200 & 697 \\
MELD & Affective Computing & Weighted F1 & 200 & 1000 \\
Flores-37/101 & Multilingual Processing & BLEU & 200 & 1012 \\ \hline
\end{tabular}
\end{table}

We split each dataset into 200 validation and $\leq 1,000$ test instances to avoid contamination. Detailed information is provided in Table \ref{tab:datasets}.

\subsubsection{Baselines}
We compare MERGEvolve against eight representative baselines:
\begin{itemize}
    \item \textbf{Best Single Expert:} The expert model that achieves the optimal performance on a specific test task. 
    \item \textbf{Data Merge:} A unified LoRA model is trained by merging fine-tuning data from 10 distinct domains. 
    \item \textbf{Expert Fusion:} A fundamental weighted-averaging strategy where weights are allocated for parameter fusion based on the normalized fitness of each expert on the target task.
    \item \textbf{TIES~\cite{yadav2023ties}:} A static merging method that mitigates parameter interference by pruning low-magnitude parameters and resolving sign conflicts.
    \item \textbf{LoraHub~\cite{huang2023lorahub}:} The merging weights are searched in the model combination space using CMA-ES.
    \item \textbf{Pack of LLMs~\cite{mavromatis2024pack}:} An information-theoretic dynamic merging scheme that dynamically allocates weights based on the perplexity of models given the input prompts.
    \item \textbf{Evolutionary Model Merge (EMM)~\cite{2025Evolutionary}:} EMM employs CMA-ES to search for layer-wise weights in the parameter space and optimizes the hierarchical combination sequences and inference paths in the data-flow space.
    \item \textbf{Model Swarms~\cite{feng2025model}:} Iteratively searches in the parameter space using PSO.
\end{itemize}

\subsubsection{Implementation Details}
We fine-tune Gemma-2-2b-it \cite{team2024gemma} into domain-specific expert models \cite{zhang2025nature} via LoRA \cite{hu2022lora}. Leveraging the Llama-Factory framework \cite{zheng-etal-2024-llamafactory}, we specialize the base model across 10 distinct domains from the Tulu-v2-SFT-mixture dataset \cite{ivison2023camels}. We adopt the official implementations and default hyperparameters for all baselines. A complete implementation of MERGEvolve and the appendix can be found at \url{https://github.com/xiaofangxd/MERGEvolve}.

For MERGEvolve, we set the hyperparameters as follows: scaling coefficient $\alpha = 0.00001$, noise standard deviation $\sigma = 0.001$, population size $N = 10$, and number of iterations $K = 10$. The utility weight $w_i$ of each expert for downstream target tasks is calculated using the following formula \cite{wierstra2014natural}:
\begin{equation}
w_i = \frac{\max\left(0, \ln\left(\frac{N}{2} + 1\right) - \log(r(i))\right)}{\sum_{j=1}^{N} \max\left(0, \ln\left(\frac{N}{2} + 1\right) - \log(j)\right)} - \frac{1}{N}, \quad i = 1, \dots, N,
\end{equation}
 where $r(i)$ denotes the rank of the $i$-th expert based on its performance ($r(i)=1$ representing the best performing expert).

\begin{sidewaystable}[htbp] 
\centering
\setlength{\tabcolsep}{2pt}
\caption{Performance comparison of MERGEvolve with other baselines on a single-task setting over five independent runs. Bold and underlined values indicate the best and second-best average results, respectively.}
\label{tab:mergevolve_results}
\resizebox{\textwidth}{!}{%
\begin{tabular}{l ccccc ccccc cc}
\toprule
\textbf{Method} & \textbf{MMLU} & \textbf{MMLUPro} & \textbf{GSM8k} & \textbf{MATH} & \textbf{MGSM} & \textbf{Flores-101} & \textbf{ARC\_C} & \textbf{CSQA} & \textbf{BBH} & \textbf{DROP} & \textbf{EmoryNLP} & \textbf{MBPP} \\
\midrule
Best Single Expert & 52.90 & 26.57 & 40.80 & 14.30 & 30.22 & 10.25 & 57.34 & 64.30 & 30.22 & 30.40 & 32.53 & 31.78 \\
Data Merge & 13.60 & 25.57 & 33.00 & 12.20 & 25.98 & 12.31 & 46.08 & 37.40 & 25.98 & 27.40 & 32.37 & 15.37 \\
TIES & 19.50 & 26.80 & 28.00 & 12.50 & 33.00 & 12.10 & 31.00 & 28.00 & 36.20 & 18.80 & \underline{34.00} & 42.00 \\
\midrule
Expert Fusion & 52.73 & 27.67 & 39.50 & 13.10 & 31.82 & 11.64 & 69.03 & 65.00 & 31.82 & 22.60 & 32.66  & 28.55 \\
LoraHub & 53.00 & 27.17 & \underline{46.47} & \underline{14.90} & \underline{34.70} & \underline{13.73} & \underline{69.97} & \textbf{66.10} & 39.30 & \underline{35.32} & 31.53 & 42.63 \\
Pack of LLMs & 53.30 & 26.50 & 37.60 & 11.96 & 30.51 & 12.43 & \textbf{70.14} & 63.30 & 39.20 & 21.56 & 31.90 & 42.50 \\
EMM & \underline{53.50} & \underline{27.84} & 38.31 & 14.16 & 30.12 & 12.56 & 69.45 & \underline{65.90} & \underline{39.35} & 25.90 & 31.18 & \underline{42.64} \\
Model Swarms & 52.74 & 26.80 & \textbf{48.90} & 14.70 & 33.90 & 11.96 & 68.60 & 63.20 & 38.58 & 31.70 & 32.70 & 42.45 \\
\midrule
\textbf{MERGEvolve} & \textbf{53.53} & \textbf{28.47} & 44.00 & \textbf{15.45} & \textbf{35.61} & \textbf{14.09} & 69.65 & 64.30 & \textbf{39.53} & \textbf{37.00} & \textbf{34.50} & \textbf{42.72} \\
\bottomrule
\end{tabular}%
}
\end{sidewaystable}

 \subsection{Experimental Results}
\label{sec5_2}

\subsubsection{Single-Task and Multi-Task Domain}

Table \ref{tab:mergevolve_results} shows the performance comparison between MERGEvolve and other baselines across 12 datasets. Our proposed MERGEvolve achieves the state-of-the-art results on 9 out of 12 datasets, with consistent improvements over the Best Single Expert in logical reasoning (+26.26\%), multilingual processing (+37.46\%), and code generation (+34.42\%).

Meanwhile, it is noted that MERGEvolve does not yield the optimal performance on three tasks, including GSM8K, ARC\_C, and CSQA. This may arise from the complex geometry of the parameter space and the high density of local optima characteristic of these specific tasks. However, MERGEvolve still exhibits a significant enhancement over the Best Single Expert, avoiding the catastrophic forgetting phenomenon typically associated with some static model merging methods. 

Table~\ref{tab:latest_results} summarizes the performance comparison across ten datasets spanning five distinct domains. MERGEvolve achieves the best results in most cases, demonstrating particularly strong performance in affective computing and logical reasoning.

\begin{table}[htbp]
\centering
\caption{Performance comparison of MERGEvolve with other representative baselines on a multi-task setting over five independent runs. Bold and underlined values indicate the best and second-best average results, respectively.}
\label{tab:latest_results}
\resizebox{\textwidth}{!}{%
\begin{tabular}{l cc cc cc cc cc}
\toprule
\multirow{2}{*}{\textbf{Method}} & \multicolumn{2}{c}{\textbf{Affective Computing}} & \multicolumn{2}{c}{\textbf{Mathematics}} & \multicolumn{2}{c}{\textbf{General Knowledge}} & \multicolumn{2}{c}{\textbf{Question Answering}} & \multicolumn{2}{c}{\textbf{Logical Reasoning}} \\
\cmidrule(lr){2-3} \cmidrule(lr){4-5} \cmidrule(lr){6-7} \cmidrule(lr){8-9} \cmidrule(lr){10-11}
& \textbf{MELD} & \textbf{EmoryNLP} & \textbf{GSM8k} & \textbf{MATH} & \textbf{MMLU} & \textbf{MMLUPro} & \textbf{ARC\_C} & \textbf{CSQA} & \textbf{DROP} & \textbf{BBH} \\
\midrule
Expert Fusion & 50.30 & 29.81 & 40.52 & 12.74 & 52.97 & 25.30 & 68.08 & 63.22 & 21.95 & \textbf{38.40} \\
LoraHub       & \underline{52.57} & 30.67 & 41.80 & 10.35 & \underline{53.12} & 26.15 & 68.87 & 63.66 & 35.22 & 38.23 \\
Pack of LLMs  & 51.70 & 29.81 & 37.90 & 11.90 & \textbf{53.40} & \underline{27.07} & 69.03 & 63.10 & 23.20 & 36.60 \\
\midrule
EMM & 51.91 & 31.07 & 41.25 & \underline{12.96} & 52.73 & \textbf{28.10} & 69.03 & \textbf{66.30} & 27.16 & \underline{38.26} \\
Model Swarms & 51.43 & \underline{31.24} & \textbf{46.85} & 12.20 & 49.60 & 26.47 & \underline{69.09} & 63.66 & \underline{36.47} & 38.20 \\
\midrule
\textbf{MERGEvolve} & \textbf{52.68} & \textbf{31.60} & \underline{42.50} & \textbf{13.10} & 52.23 & 26.63 & \textbf{70.14} & \underline{63.70} & \textbf{37.25} & \textbf{38.40} \\
\bottomrule
\end{tabular}%
}
\end{table}

\subsubsection{Generalization Ability Analysis}
To evaluate the generalization capability of MERGEvolve, we conduct zero-shot transfer experiments. Specifically, the model optimized for a source task (e.g., DROP) is assessed on an unseen target task (e.g., BBH). Table \ref{tab:GAA} presents a performance comparison between MERGEvolve and four representative baselines across typical transfer paths.

\begin{table}[htbp]
    \centering
    \caption{Cross-domain generalization performance comparison.}
    \label{tab:GAA}
    \footnotesize
    \vspace{1mm}
    \begin{tabular}{l|cc|cc|cc}
        \toprule
        \textbf{Method} & \textbf{GSM8K} $\rightarrow$ & \textbf{MATH} & \textbf{DROP} $\rightarrow$ & \textbf{BBH} & \textbf{MMLU} $\rightarrow$ & \textbf{MMLUpro} \\
        \midrule
        Pack of LLMs              & 37.57 & 11.75 & 23.63 & 37.57 & 53.26 & 26.97 \\
        LoraHub                  & \textbf{50.60} & \underline{12.60} & \underline{37.06} & \underline{38.36} & \underline{53.40} & \underline{27.03} \\
        EMM & 39.88 & 12.30 & 25.10 & 37.80 & \underline{53.40} & 26.54 \\
        Model Swarms             & \underline{48.90} & 10.25 & 31.70 & 37.80 & 52.74 & 26.92 \\
        \textbf{MERGEvolve}                & 44.00 & \textbf{13.30} & \textbf{37.50} & \textbf{38.70} & \textbf{53.53} & \textbf{27.17} \\
        \bottomrule
    \end{tabular}%
\end{table}
As shown in Table \ref{tab:GAA}, MERGEvolve achieves the best performance across all target tasks. Notably, while its initial performance on the source task GSM8K is comparatively lower, the optimized model yields superior results on the unseen MATH task. We suggest that this generalization advantage may stem from the highly templated structure of GSM8K \cite{mirzadeh2024gsm}, which often encourages shortcut learning. In contrast, MATH demands robust abstract reasoning and complex logical deduction. Overall, these results indicate that MERGEvolve effectively navigates the parameter space to mitigate template overfitting and enhance the generalized reasoning capabilities of the expert models.

\subsubsection{Ablation Studies}

MERGEvolve consists of two key components: expert model merging in the initialization stage and non-convex space exploration in the evolutionary stage. To independently assess their contributions, we conducted ablation studies on four representative benchmarks, including DROP, FLORES-101, MBPP, and MATH. We compare MERGEvolve with three variants to assess the contribution of each component:
\begin{itemize}
    \item \textbf{w/o Initialization:} Omits the initial merging phase. The evolutionary process starts from the single expert with the highest standalone performance.
    \item \textbf{w/o Utils:} Disables utility-based weight allocation during evolution. All expert models are assigned uniform weights.
    \item \textbf{w/o All:} Omits the initial merging phase and disables the utility-based weight allocation.  
\end{itemize}

\begin{table}[htbp]
\centering
\caption{Ablation study of key components in MERGEvolve. Bold indicates the best performance, and the values in brackets represent the performance degradation compared to MERGEvolve.}
\label{tab:ablation_study}
\footnotesize
\begin{tabular}{l| l|l|l|l}
\toprule
\textbf{Setting} & \textbf{DROP} & \textbf{Flores101} & \textbf{MBPP} & \textbf{MATH} \\
\midrule
\textbf{MERGEvolve} & \textbf{37.00} & \textbf{14.09} & \textbf{42.72} & \textbf{15.45} \\
w/o Initialization & 29.80 \color{green}{\scriptsize (-19.46\%)} & 4.30 \color{green}{\scriptsize (-69.48\%)} & 39.41 \color{green}{\scriptsize (-7.75\%)} & 12.90 \color{green}{\scriptsize (-16.50\%)} \\
w/o Utils & 36.37 \color{green}{\scriptsize (-1.70\%)} & 13.72 \color{green}{\scriptsize (-2.63\%)} & 42.38 \color{green}{\scriptsize (-0.79\%)} & 15.25 \color{green}{\scriptsize (-1.29\%)} \\
w/o All & 29.50 \color{green}{\scriptsize (-20.27\%)} & 8.95 \color{green}{\scriptsize (-36.48\%)} & 37.86 \color{green}{\scriptsize (-11.38\%)} & 12.50 \color{green}{\scriptsize (-19.09\%)} \\
\bottomrule
\end{tabular}
\end{table}

Based on the ablation study results in Table \ref{tab:ablation_study}, we derive two key insights:

\begin{itemize}
    \item[1)] \textbf{Importance of High-Quality Initial Point}: The merging-based initialization is critical to the performance of MERGEvolve. Removing this component (w/o Initialization) results in consistent performance degradation across all tasks. On Flores-101, performance drops by 69.48\%, underscoring that a well-initialized starting point is essential for effective search in complex parameter spaces.
    \item[2)] \textbf{Effectiveness of Utility-Based Weights}: Assigning uniform weights during evolution consistently degrades performance, indicating that treating all perturbation directions equally is suboptimal. In contrast, utility-based weights improve performance by emphasizing beneficial directions and suppressing harmful updates, thereby enabling more effective refinement.
\end{itemize}

 \subsection{Emergent Capabilities}
 \label{sec5_3}
To investigate whether the models refined by MERGEvolve surpass the capability boundaries of the initial expert population, we define the emergent capability rate (ECR). The ECR is the proportion of questions correctly answered by the final evolved model within the specific subset $\mathcal{Q}_{null}$. The subset $\mathcal{Q}_{null}$ consists of problems that all initial experts failed to solve.

\begin{figure}[htbp]
\centering
\includegraphics[width=0.5\textwidth]{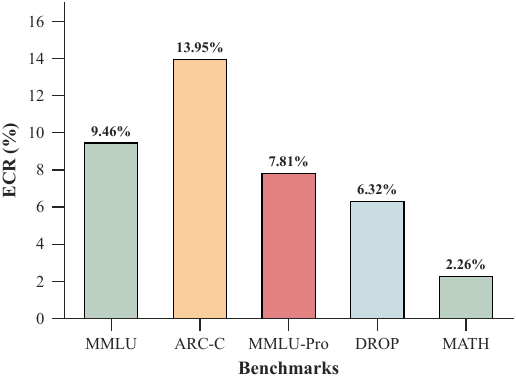}
\caption{Emergent capabilities of the evolved model by MERGEvolve across five representative benchmarks.} \label{ECR}
\end{figure}

The experimental results demonstrate that the evolved model successfully resolves a portion of complex problems that exceed the combined capacity of the initial experts. As illustrated in Fig. \ref{ECR}, the model achieves an ECR of $13.95\%$ on the ARC-C benchmark. Furthermore, the ECR reaches $9.46\%$, $7.81\%$, and $6.32\%$ on the MMLU, MMLU-Pro, and DROP benchmarks, respectively. For the highly challenging MATH benchmark, the model achieves a resolution rate of $2.26\%$ within $\mathcal{Q}_{null}$. The findings indicate that the MERGEvolve framework integrates initial expert strengths while generating novel capabilities absent from the initial population.

\subsection{Diversity of Initial Experts}
To investigate the impact of initial expert diversity on the final performance of MERGEvolve, we control the diversity levels by adjusting the number of unique expert models ($a$) and their corresponding replication counts ($b$), while maintaining a constant total population size ($a \times b = 10$), where the unique experts are selected according to their performance ranking on the target task.

\begin{figure}[htbp]
\centering
\includegraphics[width=1\textwidth]{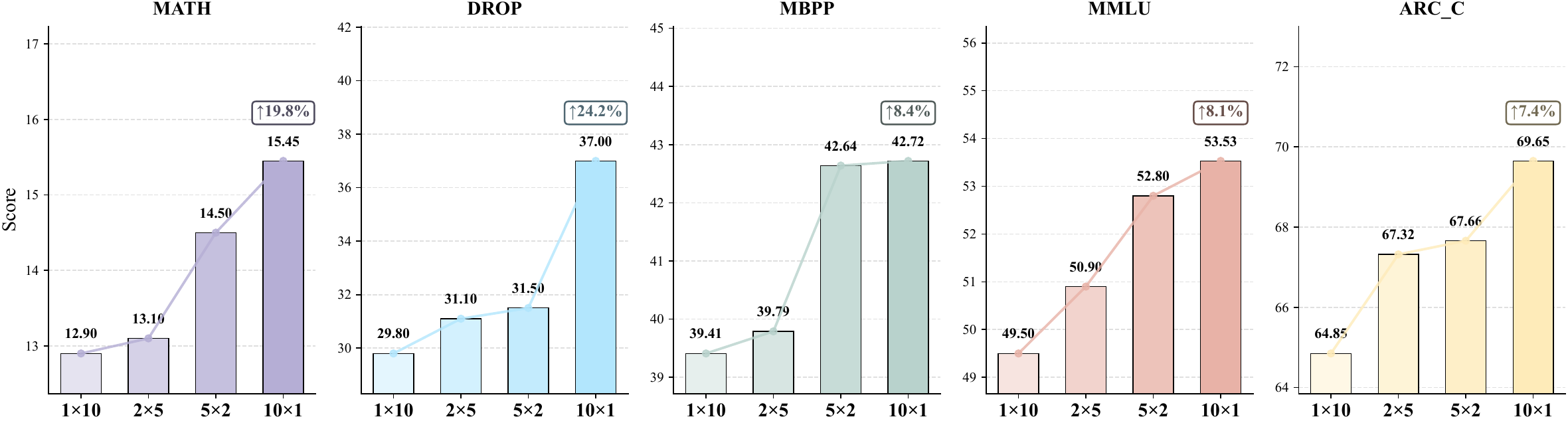}
\caption{Impact of initial expert diversity on MERGEvolve performance in five representative benchmarks.} \label{diversity}
\end{figure}

As shown in Fig. \ref{diversity}, the experimental results indicate a positive correlation between the diversity level of the initial population and the performance of the resulting evolved model. As the number of unique experts $a$ increases, the model performance across five representative benchmarks exhibits a monotonic increasing trend. The $10 \times 1$ configuration achieves an average performance improvement of 13.58\% over the $1 \times 10$ configuration. This phenomenon underscores the importance of initial expert diversity. A highly diverse set of experts provides substantial knowledge complementarity, which facilitates a more effective merging phase and broader exploration of the search space.

\section{Conclusion}
This paper presents MERGEvolve, a unified ES framework that bridges static model merging and model evolution. MERGEvolve leverages deterministic expert vectors for a high-quality initial point in the merging phase and stochastic perturbations for exploratory optimization in the evolution phase. Theoretical analysis demonstrates that MERGEvolve transcends the convex combination constraints inherent to static merging methods, enabling access to latent regions outside the restricted space. Extensive experiments on single-task and multi-task benchmarks confirm that MERGEvolve consistently achieves competitive performance against advanced baselines, while ablation studies highlight the critical role of initialization quality in facilitating efficient exploration. In the future, we plan to extend MERGEvolve to heterogeneous model architectures and develop adaptive mechanisms for dynamically adjusting perturbation scales and utility weights during evolution.

\begin{credits}
\subsubsection{Acknowledgments}
This work was supported in part by the National Natural Science Foundation of China(No.62576264) and the Natural Science Basic Research Program of Shaanxi (No. 2025JC-YBQN-795).

\subsubsection{\discintname}
The authors have no competing interests to declare that are relevant to the content of this article.
\end{credits}

%
%
\bibliographystyle{splncs04}
\bibliography{mybibliography}
\end{document}